\documentclass[10pt,twocolumn,letterpaper]{article}

\usepackage{iccv}
\usepackage{times}
\usepackage{graphicx}
\usepackage{amsmath}
\usepackage{amssymb}
\usepackage{booktabs}
\usepackage{subfig}
\usepackage{multirow}
\usepackage{algorithm}
\usepackage{algorithmicx}
\usepackage{algpseudocode}
\usepackage[bold]{hhtensor}
\usepackage[T1]{fontenc}
\graphicspath{{figures/}}

\usepackage[pagebackref=true,breaklinks=true,colorlinks,bookmarks=false]{hyperref}

\iccvfinalcopy 

\def\iccvPaperID{1544} 

\begin{document}

\title{Adaptive Feeding: Achieving Fast and Accurate Detections by Adaptively Combining Object Detectors}

\author{Hong-Yu Zhou \quad Bin-Bin Gao \quad Jianxin Wu\\
	\and
	National Key Laboratory for Novel Software Technology\\
	Nanjing University, China\\
	{\tt\small \{zhouhy,gaobb\}@lamda.nju.edu.cn, wujx2001@nju.edu.cn}
}

\maketitle

\begin{abstract}
   Object detection aims at high speed and accuracy simultaneously. However, fast models are usually less accurate, while accurate models cannot satisfy our need for speed. A fast model can be 10 times faster but 50\% less accurate than an accurate model. In this paper, we propose Adaptive Feeding (AF) to combine a fast (but less accurate) detector and an accurate (but slow) detector, by adaptively determining whether an image is easy or hard and choosing an appropriate detector for it. In practice, we build a cascade of detectors, including the AF classifier which make the easy vs. hard decision and the two detectors. The AF classifier can be tuned to obtain different tradeoff between speed and accuracy, which has negligible training time and requires no additional training data. Experimental results on the PASCAL VOC, MS COCO and Caltech Pedestrian datasets confirm that AF has the ability to achieve comparable speed as the fast detector and comparable accuracy as the accurate one at the same time. As an example, by combining the fast SSD300 with the accurate SSD500 detector, AF leads to 50\% speedup over SSD500 with the same precision on the VOC2007 test set.
\end{abstract}

\section{Introduction}
Speed and accuracy are two main directions that current object detection systems are pursuing. Fast and accurate detection systems would make autonomous cars safer, enable computers to understand scene information dynamically, and help robots act more intelligently.

The community have strived to improve both speed and accuracy of detectors. Most recent state-of-the-art detection systems are based on deep convolutional neural networks. The basic pipeline of these modern detectors can be summarized as: generate bounding box proposals, extract features for each proposal, and apply a high-quality classifier. To obtain higher accuracy, better pretrained models, improved region proposal methods, context information, and novel training strategies can be utilized. But, these methods often suffer from high computational costs, e.g., tens of thousands of region proposals are required to obtain high accuracy. On the other hand, there are a few works focusing on building faster detectors by hacking regular stages designed for traditional systems. YOLO replaces the general region proposal procedures by generating bounding boxes from regular grids~\cite{yolo}. Single Shot MultiBox Detectors (SSD) make several improvements on existing approaches, and the core of it is to calculate category scores and box offsets at a fixed set of bounding boxes using small and separated convolution kernels~\cite{ssd}. Although these approaches speed up the detection process, their accuracy rates are still lower than those slow but accurate detectors. In fact, as shown in Table~\ref{t1}, accuracy drops as speed increases.

\begin{table}[t]
	\caption{\small Speed (fps) and accuracy (mAP) of various modern detection systems. ``07+12'' means these models are trained on the combined train and validation sets of VOC07 and VOC12. The models are evaluated on the VOC07 test set. We measure the speed with batch size 1.}
	\label{t1}
	\centering
	\begin{tabular}{lcrccc}
		\toprule
		Method   & train set & FPS & mAP  \\
		\midrule
		Fast-Yolo~\cite{yolo} & 07+12 & 154 & 52.7 \\
		Yolo~\cite{yolo}      & 07+12 & 45 & 63.4 \\
		SSD300~\cite{ssd}\rlap{\textsuperscript{1}} & 07+12 & 46 & 72.1 \\
		SSD500~\cite{ssd}\rlap{\textsuperscript{1}} & 07+12 & 19 & 74.9 \\
		R-FCN~\cite{rfcn} & 07+12 & 8 & 79.0 \\
		\bottomrule
	\end{tabular}  
	\centerline{\scriptsize\rlap{\textsuperscript{1}}~\url{https://arxiv.org/pdf/1512.02325v2.pdf}.}
\end{table}

We humans are able to adaptively tune between detection speed and recognition accuracy. When you step into the kitchen, it might seem easy to find the cabinet in few milliseconds, but it surely will cost you longer to locate the toaster. However, most modern detection models ``look at'' different input images in the same way. Specifically, the time cost is nearly the same across different images. For example, regardless of the number of persons in the foreground, the region proposal network in Faster R-CNN~\cite{fasterrcnn} generates tens of thousands of proposals which will definitely decrease the processing speed in images containing only few or even no people.

In this paper, we propose to adaptively process different test images using different detection models, in which we utilize two detectors: one fast but inaccurate, and one accurate but slow. We first decide whether an input image is ``easy'' (suitable for the fast detector) or ``hard'' (for which the accurate detector is desirable), such that the test image can be adaptively fed to different detectors. We hope the entire detector to be as fast as the fast detector while maintaining the accuracy in the accurate one.

To make this promising tradeoff, we propose a novel technique, \emph{adaptive feeding} (AF), to efficiently extract features that are useful for this purpose and to learn a classifier that is simple and fast. Specifically, we build a cascade of object detectors, in which an extremely fast detector is first used to generate few instance proposals, based on which the AF classifier is able to adaptively choose either the fast or the accurate model to finish the detection task. Experiments (including timing and accuracy analyses) on several detector pairs and datasets show that there are three benefits in our AF pipeline:
\begin{itemize}
	\item The AF detector runs much faster than the accurate model (in many cases its speed is similar to or comparable to the fast model). Meanwhile, the accuracy of AF is much higher than the fast model (in many cases close to or comparable to the accurate model). Hence, by combining a fast (but inaccurate) and an accurate (but slow) model, we simultaneously achieve fast and accurate detection in AF.
	\item AF can directly utilize existing models even with different architectures. And there is no need for additional training data.
	\item AF employs an imbalanced learning framework to distinguish easy from hard images, in which it is easy to adjust the tradeoff between the speed and accuracy of the combined system.
\end{itemize}

\section{Related Work}

Object detection is one of the most fundamental challenges in computer vision, which generally consists of feature extraction at various locations (grids or proposals) and classification or bounding box regression. Prior to fast R-CNN, these two steps were usually optimized separately. Fast R-CNN~\cite{fastrcnn} employed an end-to-end learning approach to optimize the whole detector, and Faster R-CNN~\cite{fasterrcnn} further incorporated the proposal generation process into learning. Unlike these methods, we focus on the utilization of pretrained models. In this section, we review existing methods, in particular those trying to accelerate the detection.

\textbf{Detection systems.} The deformable parts model (DPM) is a classic object detection method based on mixtures of multiscale deformable part models~\cite{dpm}, which can capture significant variations in object appearances. It is trained using a discriminative procedure that only requires bounding boxes for the objects. DPM uses disjoint steps and histogram of gradients features~\cite{hog}, which is not as competitive as ConvNet-based approaches.

R-CNN~\cite{rcnn} starts another revolution of object detection after DPM. R-CNN is among the first to employ deep features into detection systems, and obtained significant improvements over existing detectors at its time. However, the resulting system is very slow because features are extracted from every object proposal. Compared with R-CNN, Fast R-CNN~\cite{fastrcnn} not only trained the very deep VGG16~\cite{vgg} network but also uses ROI pooling layer~\cite{spp} to perform feature extraction, and was 200$\times$ faster at test time. After that, to speed up the proposal generation process, Faster R-CNN~\cite{fasterrcnn}
proposed the region proposal network (RPN) to generate bounding box proposals and thus achieves improvements on both speed and accuracy. 
Recently, ResNet~\cite{resnet} begins to replace the VGG net in some detection systems, such as Faster R-CNN~\cite{fasterrcnn} and R-FCN~\cite{rfcn}. However, state-of-the-art accurate detectors are in general significantly slower than real-time.

\textbf{Fast detectors.} Accelerating the test process is a hot research topic in object detection. As rich object categories often have many variations, few research focus on the speed optimization prior to DPM. Fast detectors mainly focused on detecting a specific object, such as face and human detectors~\cite{fasthumandetector,rapidobjectdetection}. After DPM was invented, many DPM-based detectors~\cite{dpmv5,fastestdpm,singlemachine} focused on optimizing different parts in the pipeline. Dean~\etal\cite{singlemachine} exploited a locality-sensitive hashing method which achieves a mean average precision of 0.16 over the full set of 100,000 object classes. Yan~\etal\cite{fastestdpm} accelerated three prohibitive steps in the cascade version of DPM, and then get an 0.29 second average time on PASCAL VOC 2007 while maintaining nearly the same performance as DPM. Sadeghi and Forsyth~\cite{dpmv5} reimplemented the deformable parts model and achieved a near real-time version. 

In recent years, after R-CNN's invention, many works tend to speed up the detection pipeline by importing new functional layers in deep models. However, it is not until recently that we begin to approach real-time detection. YOLO~\cite{yolo} framed object detection as a regression problem to spatially separated bounding boxes and associated class probabilities, and proposed a unified architecture which is extremely fast. The SSD models~\cite{ssd} leave separated convolution kernels in charge of default proposal boxes with different size and ratios. Both YOLO and SSD share something in common: a) decrease the number of default bounding box proposals; b) employ a unified network and incorporate different stages into the same framework. These fast detectors are, however, less accurate than slow but accurate models such as R-FCN (\cf Table~\ref{t1}).

Recently, there are also some researches utilizing cascaded and boosting methods~\cite{ohem}~\cite{cascadednet}~\cite{cascadedface}~\cite{SDPCRC}~\cite{branchynet}. Shrivastava~\etal\cite{ohem} make the traditional boosting algorithm available on deep networks which achieves higher accuracy and maintain the same detection speed. Similar to ours, Angelova~\etal\cite{cascadednet} is based on sliding window and processes different image regions independently. However, recent deep detectors use fully-convolutional networks and take the whole image as input. On the contrary, our adaptive feeding method make a choice on the image level (not the region level) and thus saves a lot of time.

The proposed adaptive feeding method follows a different route: to seek a combination of two detection systems with different strengths. In this paper, we tackle such a situation: one detector is fast, the other is slow but more accurate, which widely exist as aforementioned. Our approach looks a bit like the ensemble methods because both of them rely on the \emph{diversity} of different models. However, ensemble methods often suffer from enormous computations and are difficult to implement in real-time, while our method approaches the accuracy advantage of the accurate detector and maintains the speed advantage of the fast one.

\section{Motivation: ``Easy'' vs. ``Hard'' Images} \label{easyhard}

\begin{figure*}
	\centering
	\subfloat[$P_2-P_1<0$]
	{\includegraphics[width=0.65\columnwidth]{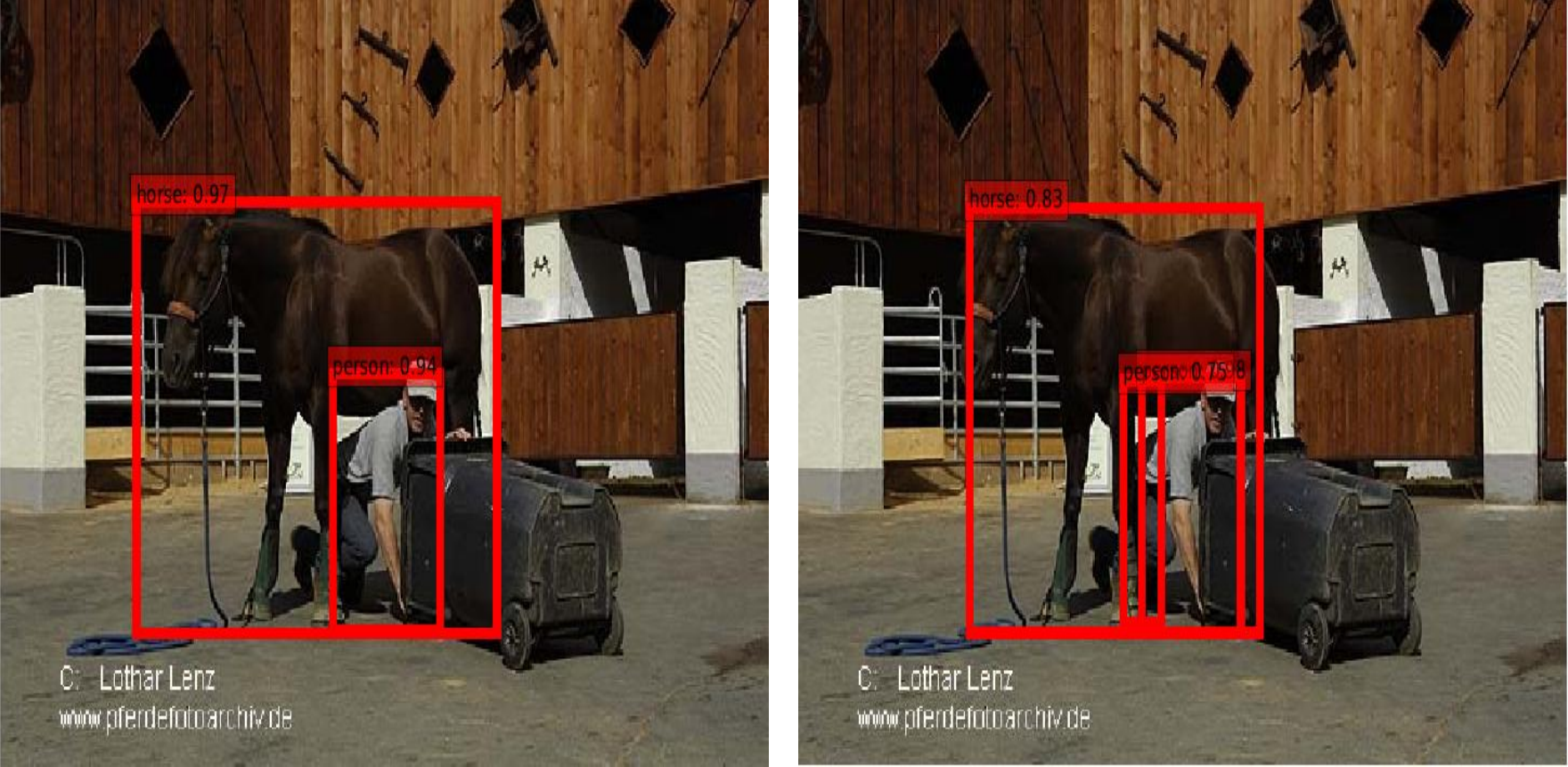} \label{fig1a}}
	\subfloat[$P_2-P_1=0$]
	{\includegraphics[width=0.65\columnwidth]{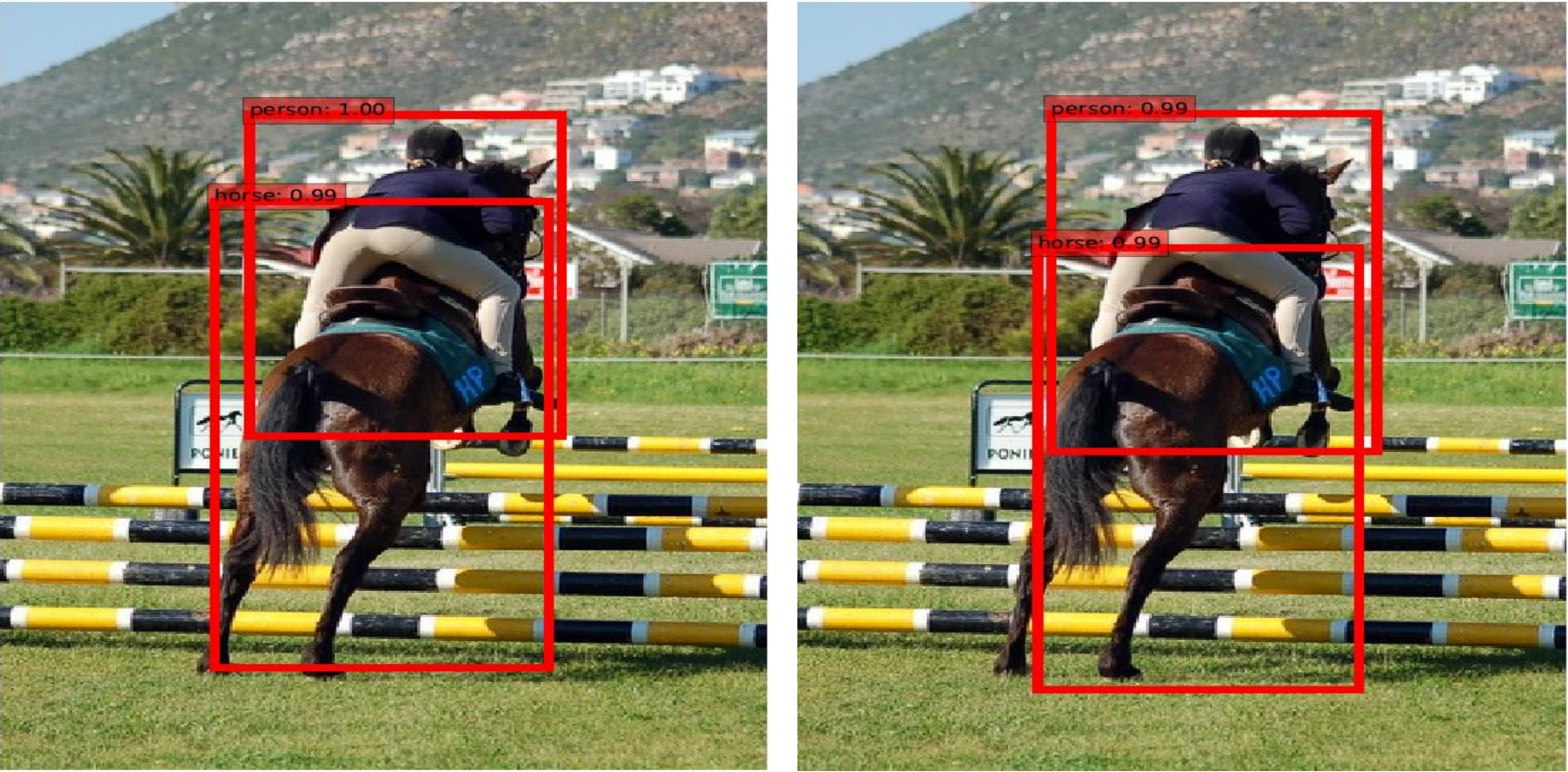} \label{fig1b}}
	\subfloat[$P_2-P_1>0$]
	{\includegraphics[width=0.65\columnwidth]{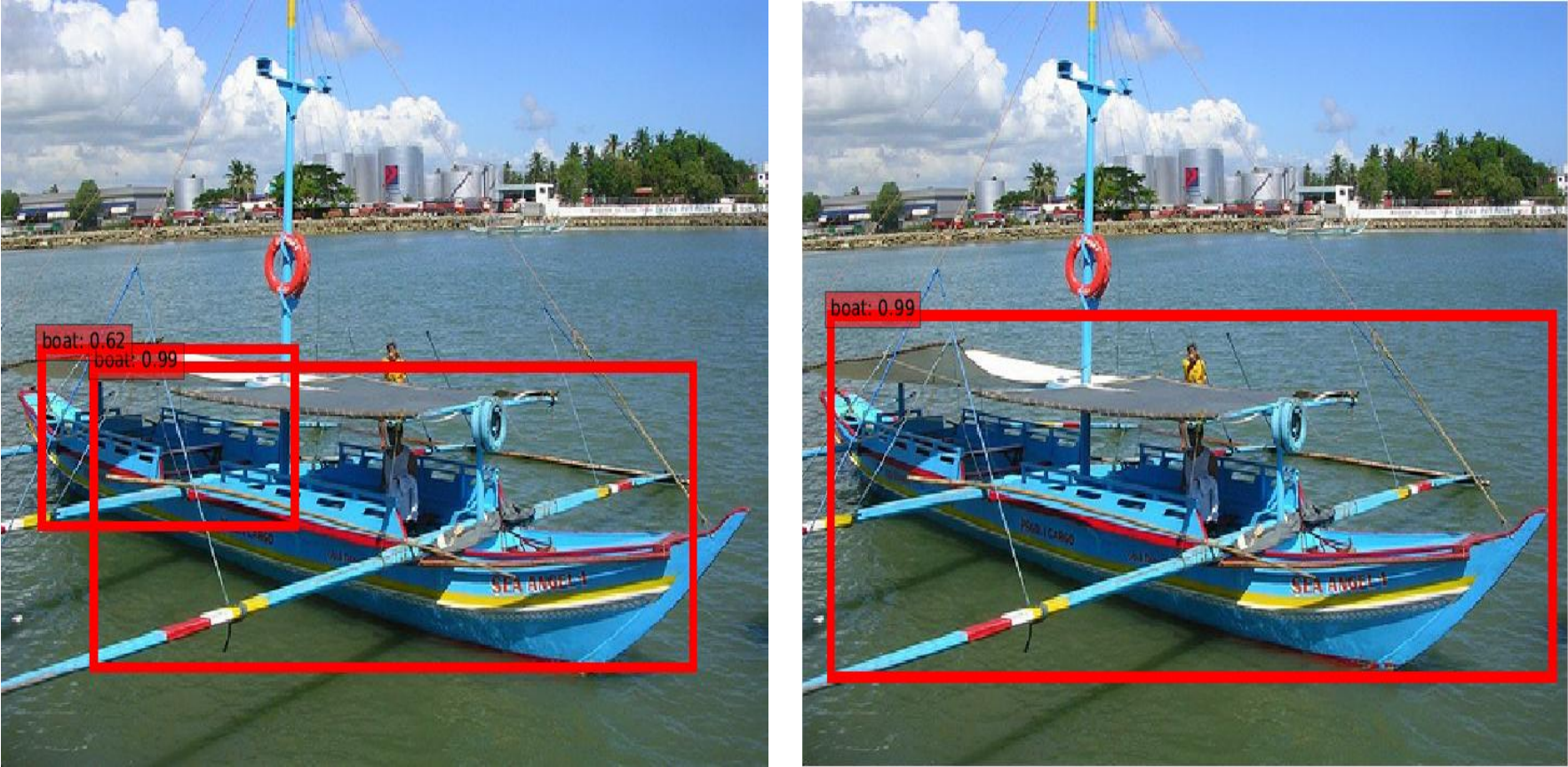} \label{fig1c}}
	\caption{\small Easy and hard images. In each figure, the left picture is the results of SSD300, and the right of R-FCN. SSD300 (the fast model) is better than, same as, or worse than R-FCN (the accurate detector) in Figure~\ref{fig1a},~\ref{fig1b},~\ref{fig1c}, respectively. $P_1$ and  $P_2$ stands for mAPI of the fast and accurate detector for the image in consideration, respectively. (Best if viewed in color.)}
	\label{f1}
\end{figure*}

Given a fast and an accurate detector, the motivation and foundation of adaptive feeding is the following empirical observation: although on average the accurate model has higher accuracy than the fast model, \emph{in most images the fast model is as precise as the accurate one} (and in few cases it is even better). For convenience, we use ``easy'' to denote these images for which the fast detector is as precise as the accurate one, and the rest the ``hard''. Furthermore, when combining these detectors, we call the fast model the \emph{basic} model, and the other more accurate detector as the \emph{partner} model. In Figure~\ref{f1}, examples are shown for cases where the basic model is better than, same as or worse than the partner model.

In order to create groundtruth labels for the easy vs. hard distinction, we apply the mean average precision (mAP) detection evaluation metric to one single image. In the PASCAL VOC Challenge~\cite{pascalvoc_v1,pascalvoc_v2}, the interpolated average precision~\cite{ap} (AP) is used to evaluate detection results. A detection system submits a bounding box for each detection, with a confidence level and a predicted class for each bounding box. For a given class, the precision/recall curve is computed from a method's ranked output based on the confidence scores. The AP metric summarizes the shape of precision/recall curve,  and a further mAP (mean average precision) averages the AP in all classes. In PASCAL VOC and MS COCO~\cite{coco}, mAP is calculated by the formula
\begin{equation}
\label{mAP}
mAP = \frac{1}{N}\sum_{i=1}^{N}{AP_i} \,,
\end{equation}
where $N$ is the number of classes in the dataset.

However, our evaluation target is a single image. Hence, we apply Equation~\ref{mAP} but focus on one image ($N=1$), as
\begin{equation}
P = \frac{1}{S}\sum_{i=1}^{S} AP_i \,, \label{mAPI}
\end{equation}
where $S$ is the number of classes \emph{in this image}, $AP_i$ is the average precision for class $i$ in this image, and $P$ represents the mean average precision in this image. In the rest of this paper, we call Equation~\ref{mAPI} mAPI, which stands for \emph{mean Average Precision per Image}.

Given two models $m_1$ and $m_2$, we assume $m_2$ is more accurate than $m_1$, but $m_1$ runs much faster than $m_2$. We evaluate both models on a set of $M$ images, which returns $\left\{P_{1,1}, P_{1,2},\dots, P_{1,M}\right\}$ and $\left\{P_{2,1}, P_{2,2}, \dots, P_{2,M}\right\}$, where $P_{i,j}$ is the mAPI for model $i$ ($i\in\{1,2\}$) and image $j$ ($1 \le j \le M$). We then split the difference set into two parts, the easy and the hard, according to a simple rule: if $P_{2,j}>P_{1,j}$ (i.e., if the accurate model has larger mAPI on image $j$ than the fast one), this image is a ``hard'' one; if $P_{2,j} \le P_{1,j}$ (i.e., if the fast model performs as good as or better than the accurate detector), this image is an ``easy'' one.

We can now collect statistics about easy vs. hard images with the groundtruth labels defined, as shown in Table~\ref{t3} for different setups. In Table~\ref{t3}, SSD300 and SSD500 are the SSD models applied to different input image sizes ($300 \times 300$ and $500 \times 500$, respectively)~\cite{ssd}. Results in Table~\ref{t3} show that \emph{most (around 80\%) images are easy}.

\begin{table}
	\centering
	\caption{\small Easy vs. hard ratios under different settings. Basic and Partner models are trained on VOC07+12 trainval.}
	\label{t3}
	\setlength{\tabcolsep}{2.5pt} 
	
	\begin{tabular}{cccccccccccc}
		\toprule
		\multirow{2}{*}{Basic} & \multirow{2}{*}{Partner} & \multirow{2}{*}{Set} & \multicolumn{2}{c}{$P_2-P_1$} \\
		\cmidrule{4-5}
		& & &$\leq$\ 0 (Easy) & \textgreater\ 0 (Hard)\\
		\midrule
		SSD300  &  SSD500 & 07+12 trainval & 83.7\% &  16.3\%   \\
		SSD300  &  SSD500 & 07 test &  81.4\% &  18.6\%   \\
		SSD300  &  R-FCN & 07+12 trainval &  89.4\% &  10.6\%   \\
		SSD300  &  R-FCN & 07 test &  78.6\% &   21.4\%   \\
		\bottomrule
	\end{tabular}
\end{table}

That is, for a large proportion (80\% or so) of \emph{easy} images, we can use $m_1$ (the fast model) for detection, which has both fast speed and accurate results; for the rest small portion (20\% or so) of \emph{hard} images, we apply $m_2$ (the slow accurate detector) to maintain high accuracy. However, since the percentage of hard examples is small, they will not significantly reduce the overall detection speed.

\section{Adaptive Feeding}

\begin{figure*}
	\centering
	{\includegraphics[width=1.8\columnwidth]{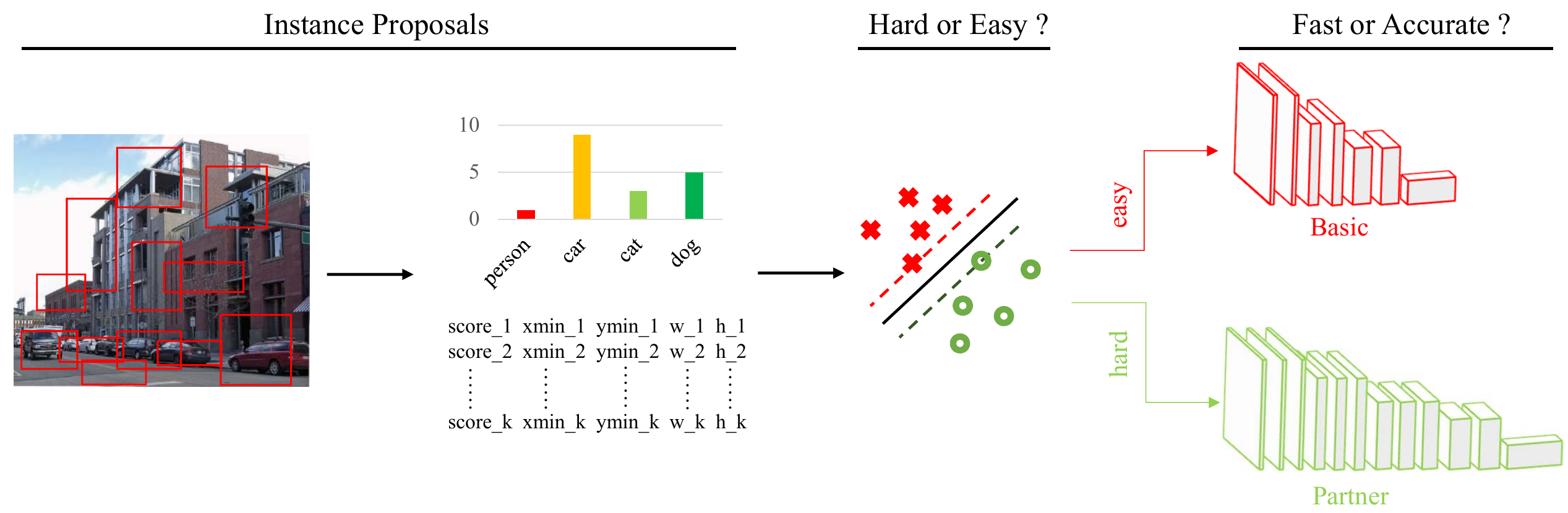} }
	\vspace{-12pt}
	\caption{\small The proposed adaptive feeding framework.}
	\label{framework}
\end{figure*}

The proposed adaptive feeding (AF) is straightforward if we know how to separate easy images from hard ones. Figure~\ref{framework} shows the framework which mainly contains three parts: instance proposals, a binary classifier and a pair of detectors. At the first step, an instance generator is employed to generate instance proposals, based on which the binary classifier decides either to feed a test image to the basic or the partner. In this section, we propose techniques for how to make this decision.

\subsection{Instance Proposals: Features for easy vs. hard}\label{af}

Since the label of ``easy'' or ``hard'' is determined by the detection results, we argue that instances in the image should play a major role. This encourages us to put an instance generator at the first stage in AF (Figure~\ref{framework}) to extract features for easy vs. hard classification. To obtain both high speed and accuracy in the following classification, we require that these proposals carry predicted class labels which will provide detailed information; and, just a few of them are able to describe the whole image well. As a result, an extremely fast detector with reasonable accuracy should be the first choice. In this paper, we use Tiny YOLO, an improved version of Fast YOLO~\cite{yolo}, as our instance generator. Tiny Yolo takes less than 5ms to process one image on a modern GPU and achieves 56.4\% mAP on VOC07, which makes the features powerful and fast to extract.

The proposals generated by the instance generator that have top confidence values contain a lot of information about objects in an image. Specifically, one proposal include three components: $C$ (predicted \underline{c}lass label), $S$ (confidence \underline{s}core) and $B$ (\underline{b}ounding box coordinates). We extract features based on the top $K$ proposals with highest confidence scores to determine whether this image is easy or hard using a binary linear support vector machine (SVM) classifier. Since the feature length is small if $K$ is small, the rest SVM classification takes little time (less than 0.1ms) per image, and is negligible.

\textbf{Ablation Analysis.} Several ways are available to organize information in $\{C,S,B\}$ into a feature vector. Ablation studies are carried out to find out the best practice using the PASCAL VOC07 dataset (which has 20 classes), with results in Table~\ref{t2}. For the simplest case, we utilize an VGG-16 model pretrained on ImageNet to do the binary classification and report the mAP in {\em{row 0}}. We can see that the basic image classification model usually has bad performance on this simple task.

The predicted class labels of $K$ proposals ($C$) can form a 20-dim histogram (denoted as ``20'' in Table~\ref{t2}), or $K$ 20-dim confidence vectors (one for each proposal, denoted as ``20-prob''). Comparing rows 3 and 6, we find that the histogram of predicted classes is not only shorter, but also more accurate. We believe the confidence for each proposal ($S$, denoted as ``conf'' in Table~\ref{t2}) is useful and it is included in all setups of Table~\ref{t2}. The $B$ information are reorganized to have two formats: ``4s'' and ``4c''. A comparison between row 2, 1 and 6 shows that removing these coordinates will reduce the mAP by at most 0.4\%, and those features including bounding box size are more powerful (comparing row 1 with row 6). Summarizing observations from these experiments, we use ``20+(conf+4s)$\times$K'' as our features. The coordinates are normalized to be within 0 and 1.

\begin{table}[t]
	\caption{\small Ablation studies about features of easy vs. hard classification. `20'': histogram of predicted classes in top $K$ proposals; ``20-prob'': predicted class probabilities for one proposal; ``conf'': confidence score for one proposal; ``4c'': (xmin, ymin, xmax, ymax), where (xmin, ymin) and (xmax, ymax) are coordinates of the top left and bottom right corners, respectively; ``4s'': (xmin, ymin, w, h), where (w, h) is the size of each proposal; ``$\times K$'': concatenate information from top $K$ proposals;;  ``Acc.'' and ``Recall'': accuracy and recall of the easy vs. hard classification. {\em{Note that SVM is trained on 07+12 trainval. And ``easy'' and ``hard'' images are balanced during the training process.}}}
	\label{t2}
	\centering
	\setlength{\tabcolsep}{3pt} 
	\small
	\begin{tabular}{*{6}{c}}
		\toprule
		\# & Feature & Acc. & Recall & 07 mAP & FPS\\
		\midrule
		{0} & raw inputs & 56.3 & 54.2 & 72.5 & -\\
		\cmidrule{1-6}
		{1} & 20+(conf+4c)$\times$25  & 74.0 & 78.9 & 74.5 & 25 \\
		\cmidrule{1-6}
		{2} & 20+(conf)$\times$25  & 72.9 & 77.2 & 74.3 & 25 \\
		\cmidrule{1-6}
		{3} & (20-prob+conf+4s)$\times$25  & 73.5 & 77.6 & 74.4 & 27 \\
		\cmidrule{1-6}
		{4} & 20+(conf+4s)$\times$10  & \textbf{76.6} & 79.7 & 74.7 & 27 \\
		\cmidrule{1-6}
		{5} & 20+(conf+4s)$\times$50  & 76.2 & 79.8 & 74.8 & 26 \\
		\cmidrule{1-6}
		{6} & 20+(conf+4s)$\times$25  & 75.3 & \textbf{80.0} & \textbf{75.0} & \textbf{27} \\
		\bottomrule
	\end{tabular}
\end{table}

We also evaluated the effect of $K$. Comparing rows 4, 5 and 6, we find that too many proposals ($K=50$) not only reduces speed, but also lowers accuracy. Too few ($K=10$) proposal also lead to lower accuracy. Hence, we choose $K=25$ to extract our feature vector on Pascal VOC dataset, which is the last row in Table~\ref{t2}.

\subsection{Learning the easy vs. hard classifier} \label{sec:classifier}

It is not trivial to learn the easy vs. hard SVM classifier, even after we have fixed the feature representation. This classification problem is both imbalanced~\cite{me:Liu2009} and cost-sensitive~\cite{cost}, whose training requires special care.

As shown in Table~\ref{t3}, most images are easy (i.e., suitable for the fast detector). Hence, a classifier with low error rate may make a lot of errors in the hard images. For example, if the classifier simply classifies all images as easy, its classification accuracy is around 80\% (could even be as high as 89.4\%, \cf Table~\ref{t3}). However, this simple classifier will reduce AF to the fast detector, whose detections are not accurate enough.

Beyond the classification accuracy, we also want the classifier to correctly predict a high percentage of hard examples. This requires a high \emph{recall}, which is the percentage of positive examples ($y=1$, i.e., hard images) to be correctly classified. A high recall ensures that the slow accurate detector is applied to appropriate examples such that the AF detection results will be accurate.

A simple approach to solve the class imbalance problem is to assign different resampling weights for hard and easy images. Because we use SVM as our easy vs. hard classifier, this can be equivalently achieved by assigning different misclassification costs for hard and easy images. Suppose we have a set of training examples $(\vec{x}_i,y_i)$ ($1 \le i \le n$), where $\vec{x}_i$ is the AF features extracted for the $i$-th image, and $y_i \in \{-1,1\}$ is its label (easy or hard). A linear classifier $\mathrm{sgn}(\vec{w}^T\vec{x}+b)$ is learned by solving the following standard SVM problem:
\begin{align}
	\min_{\vec{w},b} & \quad \frac{1}{2}\vec{w}^T \vec{w} + C \sum_{i=1}^n \xi_i \\
	\mathrm{s.t.} & \quad y_i \left(\vec{w}^T \vec{x}_i + b\right) \ge 1 - \xi_i, \xi_i \ge 0, \, 1 \le i \le n \,,
\end{align}
in which $C>0$ is a hyperparameter that balances between large margin and small empirical error, and $\xi_i$ is the cost associated with the $i$-th image $\vec{x}_i$.

In this standard SVM formulation, easy and hard images are treated equally. In order to obtain high recall, a classic method is to assign different weights for different images. We fix the resampling weight for easy images as $c_{-1}=1$ and the resampling weight $c_{+1}>1$ for hard images~\cite{cost}. A larger $c_{+1}$ value puts more emphasis on the correct classification of hard images, and hence will in general lead to higher recall. The SVM problem is now
\begin{align}
	\min_{\vec{w},b} & \quad \frac{1}{2}\vec{w}^T \vec{w} + C \sum_{i=1}^n c_{y_i} \xi_i \\
	\mathrm{s.t.} & \quad y_i \left(\vec{w}^T \vec{x}_i + b\right) \ge 1 - \xi_i, \xi_i \ge 0, \, 1 \le i \le n \,.
\end{align}

\begin{figure}
	\centering
	\subfloat[SSD300 vs. SSD500]
	{\includegraphics[width=0.5\columnwidth]{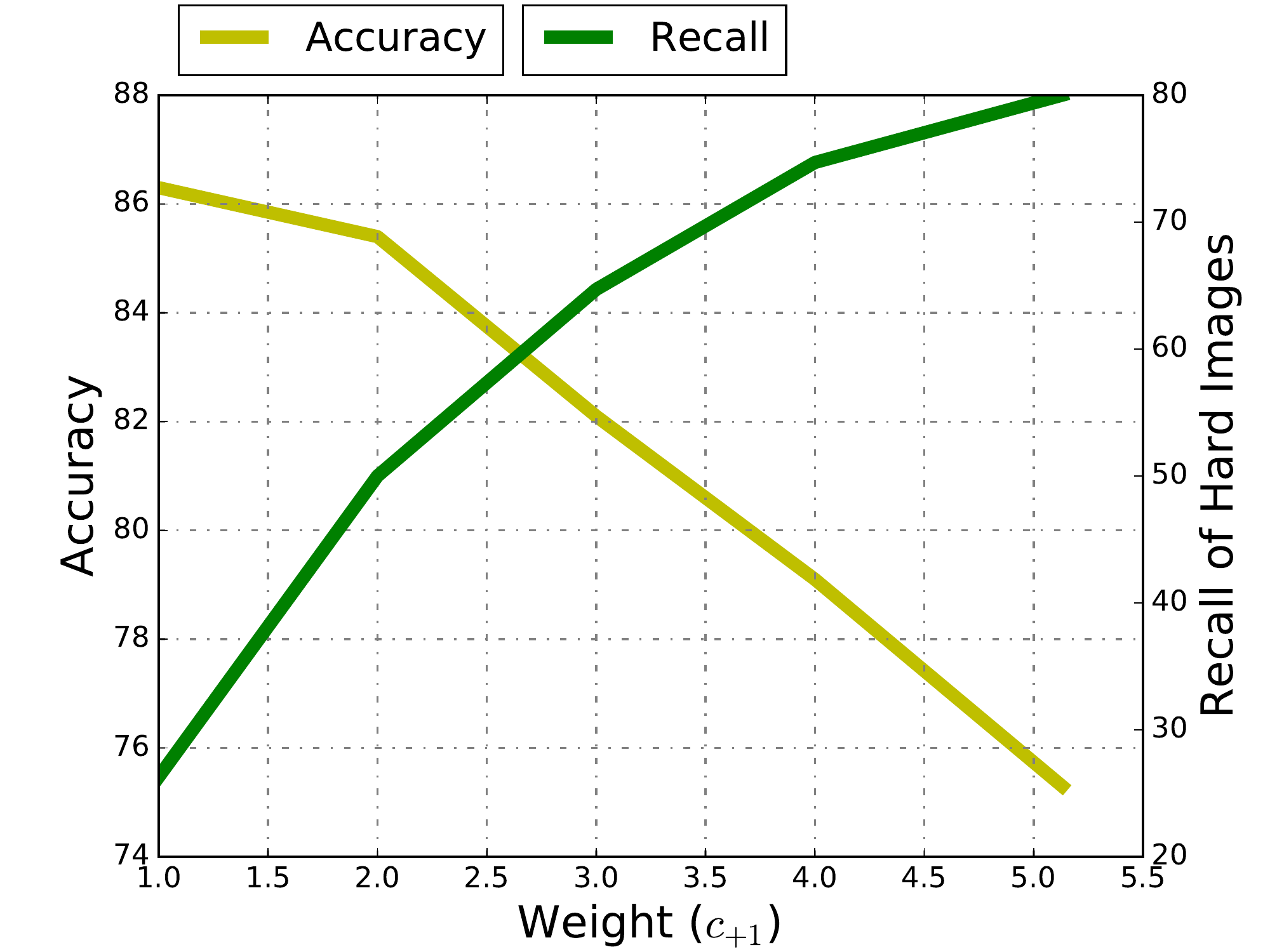}\label{w3}} 
	\subfloat[SSD300 vs. R-FCN]
	{\includegraphics[width=0.5\columnwidth]{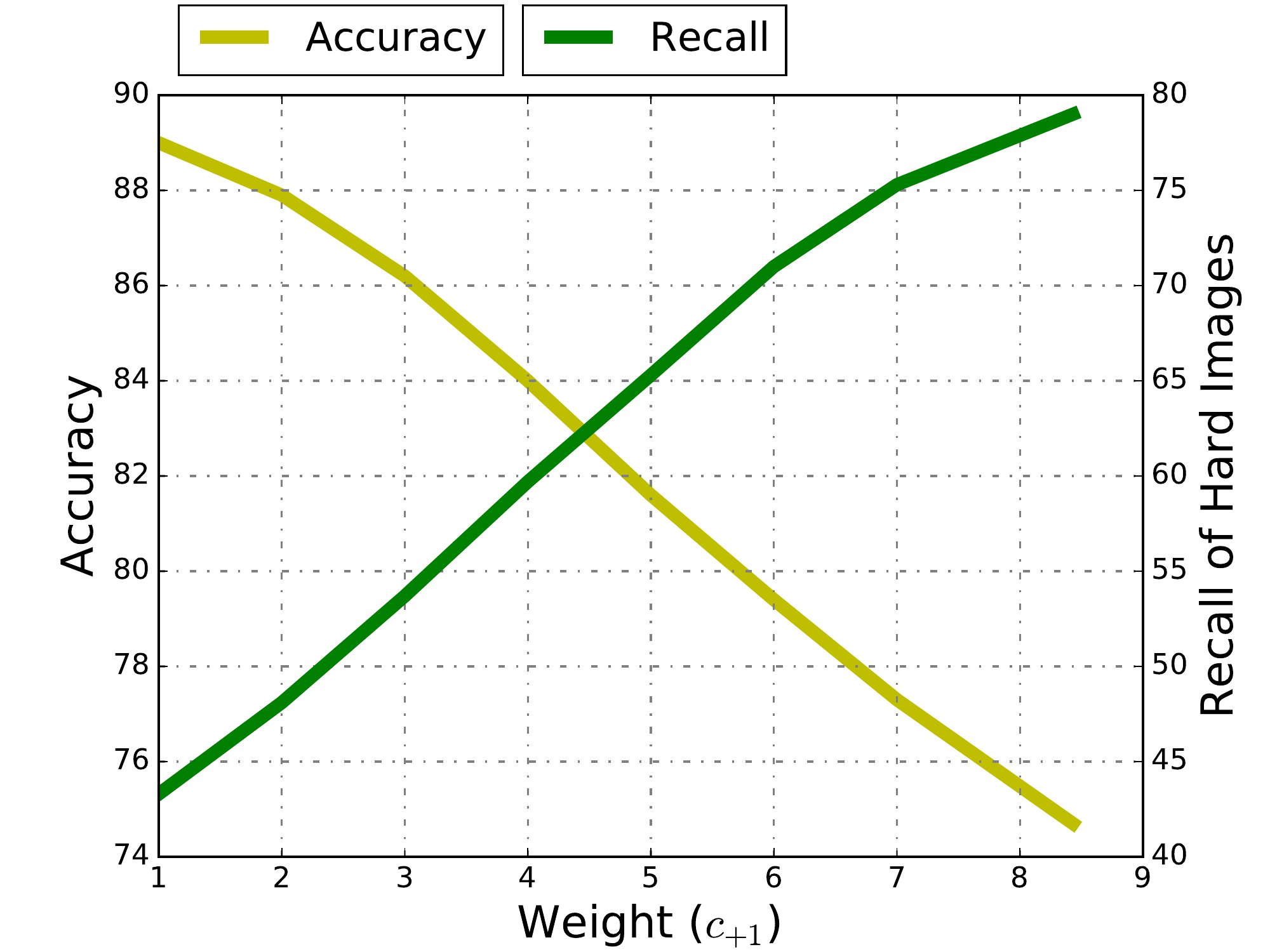}\label{w4}}
	\caption{\small Impact of sampling weights on accuracy and recall of hard images of easy vs. hard classification. The experiments are performed on VOC07 test.}
	\label{f3}
\end{figure}

\begin{figure}
	\centering
	\subfloat[SSD300 vs. SSD500]
	{\includegraphics[width=0.5\columnwidth]{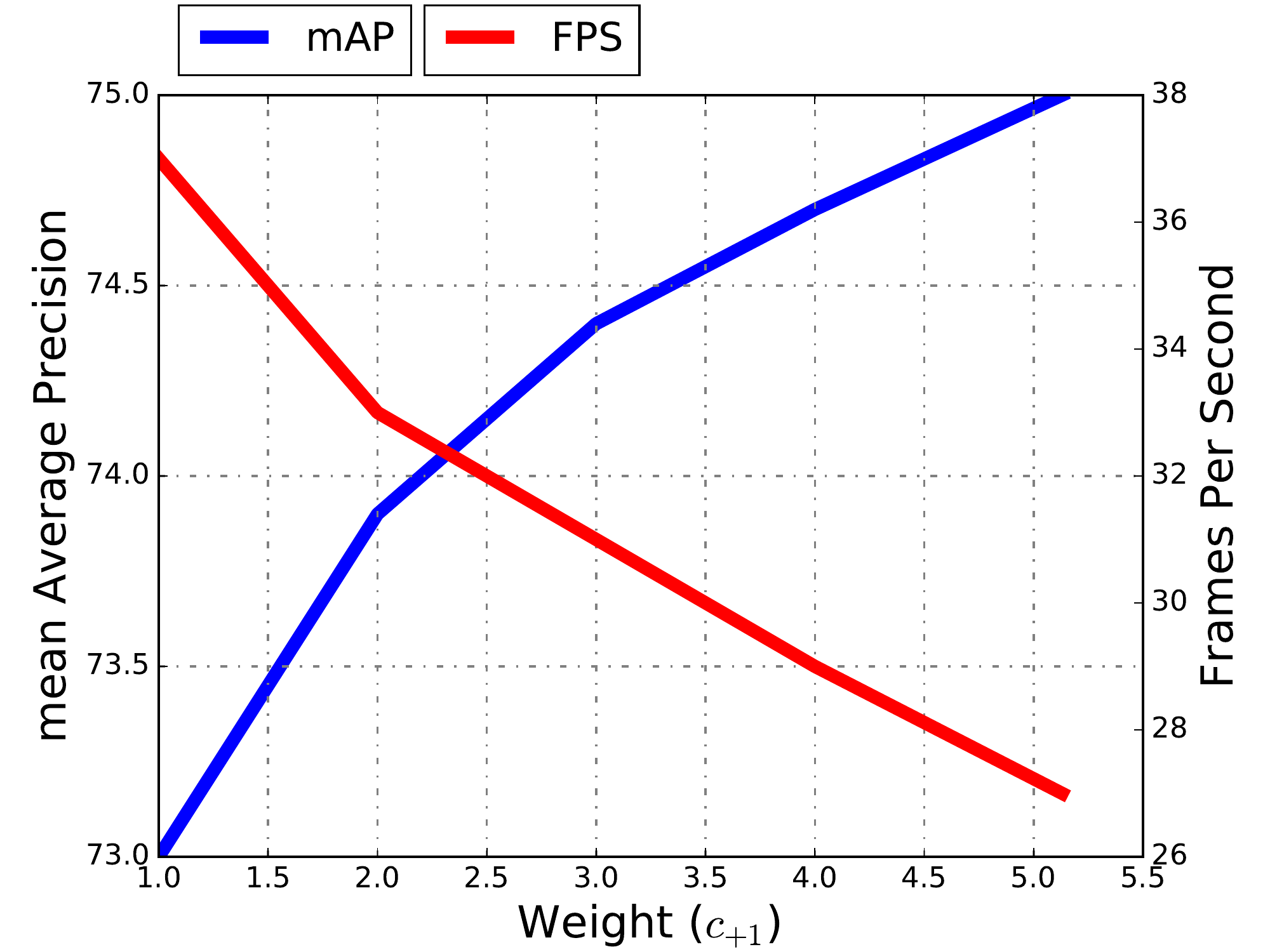}\label{w1}} 
	\subfloat[SSD300 vs. R-FCN]
	{\includegraphics[width=0.5\columnwidth]{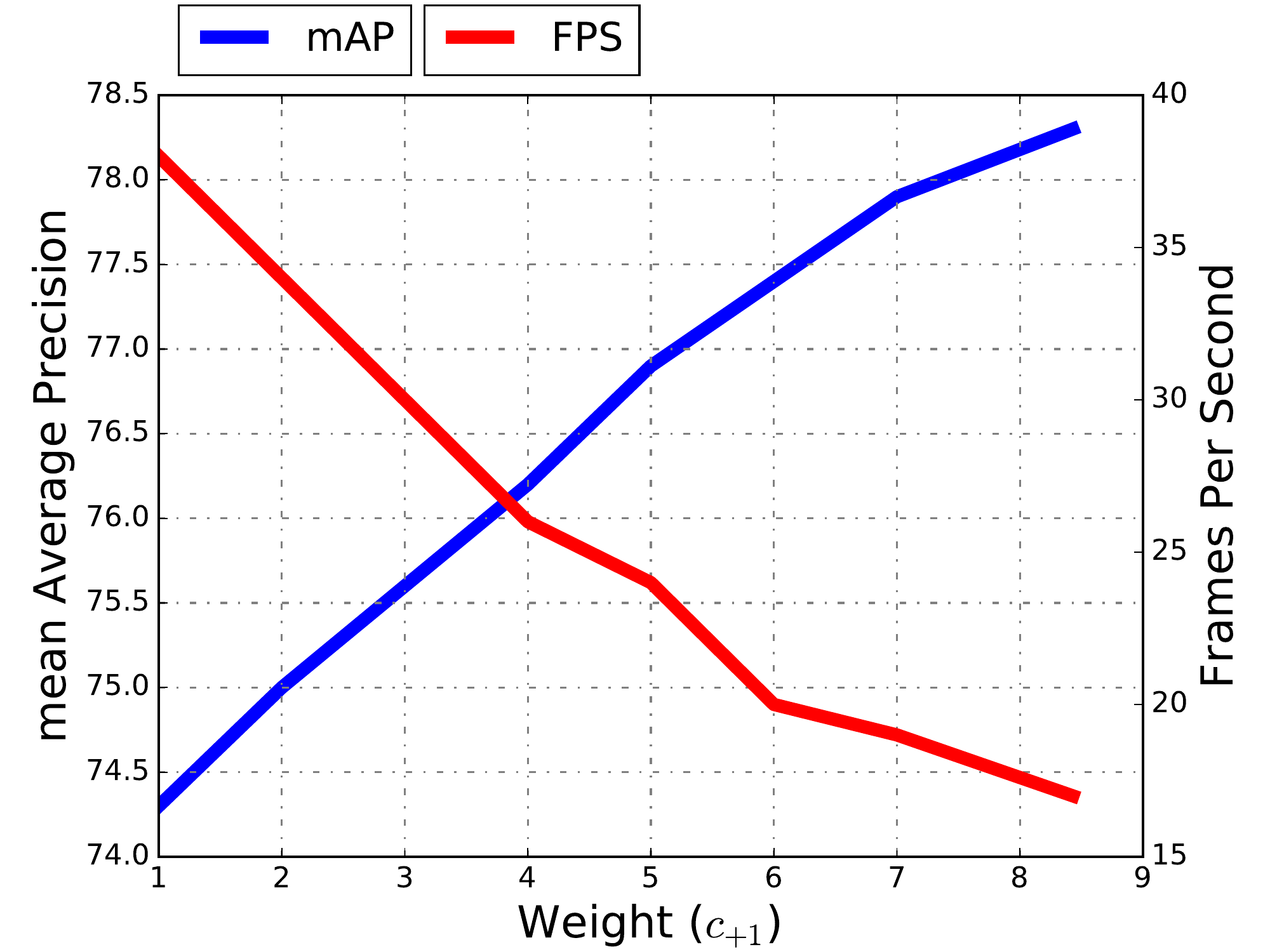}\label{w2}}
	\caption{\small Impact of sampling weights on mAP and FPS. The experiments are performed on VOC07 test.}
	\label{f4}
\end{figure}

\textbf{Ablation Analysis}. We start $c_{+1}$ from the balanced resampling ratio (i.e., ratio between the number of easy and hard images), and gradually decrease it. In the pair of SSD300 vs. SSD500, as shown in Figure~\ref{w3} and Figure~\ref{w1}, treating easy and hard examples equally ($c_{+1}=1$) leads to a low recall rate and low detection mAP, but the detection speed is very fast. When $c_{+1}$ gradually increases, the classification accuracy and fps gradually decrease but the recall rate keeps increasing. Accordingly, the detection becomes more accurate, but at the price of dropped detection speed. The same trends hold for SSD300 vs. R-FCN, too, as shown in Figure~\ref{w4} and Figure~\ref{w2}.

We note that it is a practical method to adjust the tradeoff between detection speed and accuracy by adjusting the resampling weight $c_{+1}$. When you care more about the precision, a balanced weight could be the first choice, otherwise a lower weight might fit the situation. However, in both cases, our AF achieves considerable speed-up ratio.

\section{Experimental Results on Object Detection}

We evaluate our method on the VOC 2007 test set, VOC 2012 test set~\cite{pascalvoc_v2} as well as MS COCO~\cite{coco}. We demonstrate the effect on achieving fast and accurate detections when combining two models using our AF approach.

\subsection{Setup}

We implement the SVM using scikit-learn~\cite{sklearn} and set $C=1$. We use the LIBLINEAR solver in the primal space. We use the default mode in scikit-learn for setting the resampling weight. For the basic and partner models, we directly use those publicly available pretrained detection models without any change if not specifically mentioned. All evaluations were carried out on an Nvidia Titan X GPU card, using the Caffe deep learning framework~\cite{caffe}.

For all experiments listed below, we utilize the Tiny YOLO detector as the instance generator if not otherwise specified. We choose Tiny YOLO because it is one of the fastest deep learning based general object detectors. SSD300 is used as the basic model because it runs fast and performs well. On the other side, SSD500 and R-FCN are the partner models in different experiments, because their detections are more accurate than SSD300.

On the PASCAL VOC datasets, these models are trained on VOC07+12 trainval, and tested on both VOC07 test and VOC12 test. For the sake of fairness, we don't train with extra data (VOC07 test) when testing on VOC12 test. We also conduct experiments on MS COCO~\cite{coco} and report numbers from the test-dev 2015 evaluation server.

\subsection{PASCAL VOC 2007 results}

On this dataset, we perform experiments on two pairs of models: SSD300 vs. SSD500 and SSD300 vs. R-FCN, and compare against SSD300, SSD500 and R-FCN. Specifically, the training data is VOC07+12 trainval (16551 images) and test set
is VOC07 test (4952 images). Experimental results are displayed in Table~\ref{t4}. Since we do not have an independent validation set, we train the SVM on VOC07+12 trainval. We randomly split the 16551 images into two parts, where the first part contains 13,000 images and the second keeps the rest 3,551 images. We train our SVM on the 13k set and validate it on the 3.5k images. We use 20+(conf+4s)$\times$25 as the features for SVM learning because it performs the best among different types of features in Table~\ref{t2}.

\begin{table}
	\caption{\small VOC 2007 test set detection mAP (\%). All detectors and the instance generator are trained on VOC07+12 trainval. SVM is trained on VOC07+12 trainval. The Speed-Up Ratio (SUR) and Decreased mAP (DmAP) are all based on partner model. \textbf{A}: the accurate mode. \textbf{F}: the fast mode. \textbf{W}: the sampling weight of easy to hard when training SVM.}
	\label{t4}
	\centering
	\setlength{\tabcolsep}{5pt} 
	\small
	\begin{tabular}{lrcrrr}
		\toprule
		\textbf{Method} & \textbf{W} & \textbf{mAP} & \textbf{FPS} & \textbf{SUR} & \textbf{DmAP}\\
		\midrule
		SSD300 & - & 72.1 & 46 & - & -\\
		SSD400 & - & 74.0 & 32 & - & -\\
		SSD500 & - & 74.9 & 19 & - & -\\
		Simple Ensemble & - & 73.0 & 19 & - & -\\
		R-FCN & - & 79.0 & 8 & - & -\\
		\toprule
		300-500-A & 5.13 & 75.0 & {27} & 42\% & -0.1 \\
		300-500-F & 3 & 74.4 & {33} & 74\% & 0.5 \\
		\toprule
		300-R-FCN-A & 8.43 & 78.3 & {17} & 113\% & 0.7\\ 
		300-R-FCN-F & 5 & 76.9 & {24} & 200\% & 2.1\\
		\bottomrule
	\end{tabular}
\end{table}

We provide two modes during the combination: accurate (A) or fast (F). The accurate mode takes a balanced sampling weight (5.13 for SSD500, and 8.43 for R-FCN), while the fast mode uses a lower weight (3 and 5, respectively). Compared with SSD500, 300-500-A has a slightly higher performance because the classifier makes the right choice for those images fit for the basic model. 300-R-FCN-F even outperforms SSD500 by two percent points while runs 5fps faster. If we compare AF with R-FCN, 300-R-FCN-A achieves 113\% speed-up ratio at a slight cost in mAP (0.7 points). As additional baselines, we also make experiments on SSD400 and a simple ensemble of SSD300 and SSD500. We implement SSD400 following the instructions in~\cite{ssd}. 300-500-F surpass SSD400 by 0.4 points while reaches the same speed with negligible training cost. The Simple Ensemble method brutely combines the detection results of SSD300 and SSD500 but its mAP is worse than SSD500.

\begin{table*}
	\scriptsize
	\caption{\small VOC 2012 test detection AP (\%), mAP and speed (in FPS). All detectors and the instance generator are trained on VOC07+12 trainval.}
	\label{t5}
	\centering
	\footnotesize
	\setlength{\tabcolsep}{1.8pt} 
	\begin{tabular}{l r r r r *{20}{c}}
		\toprule
		\textbf{Method} & \textbf{W} & \textbf{mAP} & \textbf{FPS} && aero & bike & bird & boat & bottle & bus & car & cat & chair & cow & table & dog & horse & mbike & person & plant & sheep & sofa & train & tv\\
		\toprule
		Fast R-CNN & - & 70.0  & 7 && 77.0 & 78.1 & 69.3 & 59.4 & 38.3 & 81.6 & 78.6 & 86.7 & 42.8 & 78.8 & 68.9 & 84.7 & 82.0 & 76.6 & 69.9 & 31.8 & 70.1 & 74.8 & 80.4 & 70.4 \\ 
		R-FCN & - & 74.9 & 8 && 84.5 & 80.4 & 77.6 & 64.0 & 60.1 & 79.2 & 78.9 & 91.5 & 55.8 & 78.3 & 57.8 & 89.6 & 84.0 & 83.7 & 83.0 & 54.5 & 79.0 & 65.1 & 83.2 & 69.3 \\
		SSD300 & - &  68.9 & 46 && 82.7 & 75.3 & 68.8 & 52.7 & 40.5 & 78.7 & 72.4 & 87.3 & 48.2 & 72.1 & 58.4 & 84.0 & 79.1 & 80.2 & 75.9 & 41.7 & 70.6 & 66.1 & 80.3 & 63.3 \\
		SSD500 & - & 71.1 & 19 && 83.1 &  77.4 & 72.4 & 54.7 & 48.2 & 78.5 & 77.1 & 87.5 & 51.5 & 75.0 & 56.1 & 84.9 & 82.2 & 81.8 & 79.7 & 45.2 & 75.4 & 63.8 & 81.8 & 66.1 \\
		\toprule
		300-500-A & 5.13 & 71.1 & 27 && 83.9 & 77.7 & 71.3 & 54.7& 46.8 & 78.7 & 77.0 & 87.6 & 51.7 & 75.4 & 57.0 & 85.1 & 82.4 & 82.1 & 79.3 & 44.3 & 74.9 & 65.6 & 81.3 & 65.3 \\
		300-500-F & 3 & 70.8 & 33 && 83.9 & 77.0 & 70.1 & 54.3 & 46.0 & 79.0 & 76.2 & 87.3 & 51.2 & 74.7 & 58.0 & 84.7 & 82.0 & 81.7 & 78.7 & 44.4 & 74.3 & 65.5 & 80.8 & 65.6 \\
		300-R-FCN-A & 8.43 & 73.9 & 17 && 84.8 & 80.4 & 73.6 & 62.0 & 54.2 & 79.7 & 78.9 & 88.7 & 53.2 & 77.4 & 58.0 & 86.7 & 84.0 & 85.3 & 82.4 & 51.8& 78.3 & 66.9 & 83.2 & 68.3 \\
		300-R-FCN-F & 5 & 73.0 & 24 && 84.0 & 78.9 & 72.0 & 60.7 & 51.0 & 79.2 & 78.1& 88.2 & 52.2 & 76.4 & 58.4 & 86.1 & 83.8 & 84.6 & 81.5 & 50.9 & 77.6 & 67.2 & 82.9 & 65.8 \\
		\bottomrule
	\end{tabular}
\end{table*}

\subsection{PASCAL VOC 2012 results}

The same two pairs of detectors are evaluated on the VOC12 test set. Accurate and fast modes are both performed, respectively. For consideration of consistence, we take the same group of sampling weights for all four combinations: 5.13 for 300-500-A, 3.0 for 300-500-F, 8.43 for 300-R-FCN-A, and 5.0 for 300-R-FCN-F. Similar to VOC 2007 test, the SVM classifier is also trained on instance proposals from VOC07+12 trainval.

Table~\ref{t5} shows the results on VOC 2012 test. The AF approach shows different effects in different pairs. For the accurate mode, 300-500-A improves the mAP from 68.9\% (SSD300) to 71.1\% which is the same as SSD500, but is 8fps faster than SSD500 (about 42\% speed-up ratio). Even though its speed (27fps) is slower than the basic model (SSD300, 46fps), its speed is still faster than what is required for real-time processing. 300-R-FCN-A runs twice as fast as R-FCN, while only loses 1.0 mAP. For the fast mode, 300-500-F runs much faster while keeps comparable precision with SSD500. 300-R-FCN-F not only performs 2.0 points higher but also is 5fps faster when compared with SSD500.

\subsection{MS COCO results}

To further validate our approach, we train and test our approach on the MS COCO dataset~\cite{coco}. All models are trained on trainval35k~\cite{ion} (including Tiny YOLO). For convenience, we use minival2014 (about 5,000 images) to train the SVM. Note that minival2014 is not contained in trainval35k. Since there are 80 classes in MS COCO, we take the top 50 proposals and the feature is (80+(conf+4s)$\times$50) for SVM learning.  This datasets exhibits different property than the PASCAL VOC datasets. From Table~\ref{t6}, in both pairs, the ratio of easy to hard images is only slightly larger than 1, which can be explained by the fact that MS COCO is ``harder'' than PASCAL VOC, because there are many small objects in it. However, although this ratio is not as large as that in VOC datasets, the adaptive feeding method still achieves convincing results.

\begin{table}
	\centering
	\caption{\small Statistics on COCO minival2014.}
	\label{t6}
	\setlength{\tabcolsep}{2.5pt} 
	\small
	\begin{tabular}{cccccccccc}
		\toprule
		\multirow{2}{*}{Basic} & \multirow{2}{*}{Partner} & \multirow{2}{*}{Set} & \multicolumn{2}{c}{P2-P1} \\
		\cmidrule{4-5}
		& & &$\leq$\ 0 (Easy) & \textgreater\ 0 (Hard)\\
		\midrule
		SSD300 &  SSD500 & minival2014 & 53.6\% &  46.4\%   \\
		SSD300 & R-FCN & minival2014 & 51.7\% & 48.3\% \\
		\bottomrule
	\end{tabular}
\end{table}

Table~\ref{t7} shows the AF results on MS COCO test-dev 2015. For the accurate mode, on the standard COCO evaluation metric, SSD300 scores 20.8\% AP, and our approach improves it to 23.7\%. 
It is also interesting to note that, since SSD500 and R-FCN are better at small and medium sized objects, our approach improves SSD300 mainly on these two parts.

\begin{table}
	\caption{\small MS COCO 2015 test-dev detection AP (\%). The SSD results are from~\cite{ssd}. R-FCN is trained by ourselves because the model in~\cite{rfcn} hasn't been released yet (slightly lower results than the official ones).}
	\label{t7}
	\footnotesize
	\centering
	\setlength{\tabcolsep}{1.8pt} 
	\begin{tabular}{l r *{8}{c}}
		\toprule
		\textbf{Method} & \textbf{W} & \textbf{test} &\textbf{FPS} & AP & AP$^{50}$ & AP$^{75}$ &AP$^{S}$ & AP$^{M}$ & AP$^{L}$ \\
		\toprule
		SSD300 & - & test-dev & 46 & 20.8 & 38.0 & 20.5 & 3.9 & 18.5 & 38.7 \\
		SSD500 & - & test-dev & 19 & 24.4 & 43.7 & 24.7 & 7.2 & 25.3 & 40.1 \\
		R-FCN & - & test-dev & 8 & 28.6 & 48.8 & 30.1 & 8.8 & 31.4 & 44.1 \\
		\toprule
		300-500-A & 1.16 & test-dev & 25 & 23.7 & 42.7 & 23.9 & 6.6 & 23.7 & 39.8 \\
		300-500-F & 2 & test-dev & 31 & 23.0 & 41.6 & 23.1 & 6.0 & 22.2 & 39.5 \\
		300-R-FCN-A & 1.07 & test-dev & 15 & 27.0 & 47.4 & 29.2 & 7.5 & 29.9 & 43.1 \\
		300-R-FCN-F & 2 & test-dev & 21 & 26.2 & 46.7 & 27.9 & 6.3 & 28.8 & 41.9 \\ 
		\bottomrule
	\end{tabular}
\end{table}

\subsection{Feature Visualization}
In this part, we want to find what makes an input image an ``easy'' or ``hard'' one.
To achieve this goal, a visualization of the learned SVM model is needed. We use linear SVM and set +1 for hard while -1 for easy ones. Features are learned on VOC07+12 trainval and COCO minival, respectively. We only report results for the natural balancing weights, and it is worth noting that there can be small changes when employing different class weights.

\begin{table}
	\caption{We categorize features into groups, which are defined in Table~\ref{t2}. The table below shows the sum of weights in each group (there are 20 and 80 classes in VOC and COCO, respectively). For better comparison, we normalize the sum of weights to 1.}
	\label{t9}
	\normalsize
	\centering
	\setlength{\tabcolsep}{1.8pt} 
	\begin{tabular}{lccccccc}
		\toprule
		{Method} & {dataset} & {class}& {conf} & {xmin} & {ymin} & {width} & {height}\\
		\midrule
		300-500 & voc & 0.22 & 1.46 & 0.08 & 0.02 & -0.55 & -0.23 \\
		300-R-FCN & voc & 0.25 & 1.58 & 0.13 & 0.05 & -0.89 & -0.12 \\
		\midrule
		300-500 & coco & 0.31 & 1.46 & -0.02 & -0.05 & -0.37 & -0.33 \\
		300-R-FCN & coco & 0.36 & 1.48 & -0.05 & -0.01 & -0.60 & -0.18 \\
		\bottomrule
	\end{tabular}
\end{table}

From Table~\ref{t9}, we can see that "conf" (confidence in top-k proposals) is the most influential factor. Thus, many high-confidence proposals make the input image a "hard" one. This fits our intuition: an image with many objects might be hard for a detector. 

There are some other interesting observations 1) large proposal hints "easy" images. We argue that images with large objects often have few instances, especially in VOC and COCO; 2) "easy" images prefer shorter proposals (w>h) while "hard" images like taller instances; 3) xmin and ymin have small weights, hence positions of proposals have small impact.

\section{Pedestrian Detection}
Since pedestrian detection can be regarded as a special case of object detection, we also apply our adaptive feeding approach to existing detectors on the Caltech Pedestrian dataset.

The basic model employs a stand-alone region proposal network (RPN). The original RPN in Faster R-CNN~\cite{fasterrcnn} is developed as a class-agnostic detector with multi-scale anchors to describe different ratios of objects at each position. Zhang~\etal\cite{fasterpd} found that a single RPN also comes with good performance in pedestrian detection. In our experiments, we build RPN based on VGG-16 and follow the strategies in~\cite{fasterpd} when designing anchors on the Caltech Pedestrian dataset. For the partner model, we directly use CompACT~\cite{compactdeep}, which proposed a novel algorithm for learning a complexity-aware cascade. In our case, we make use of CompACT-Deep which incorporates CNN features into the cascading detectors.

Note that RPN achieves an MR (missing rate) of 15.2\% on the Caltech Pedestrian testset at 10fps. CompACT-Deep reaches 12.0\% MR, but is 7fps slower than RPN. The easy to hard ratio between these two detectors is 4.25, which seems to be a good situation for AF. RPN is also used as an instance generator here, which means every input image should first pass the RPN to make a decision. For the feature inputs to SVM, we employ a similar format: ((conf+4s)$\times$25). The settings of linear SVM are the same as those in object detection.

\begin{table}
	\caption{\small MR (\%) of AF on Caltech Pedestrian dataset. The Speed-Up Ratio (SUR) and Decreased MR (DMR) are all based on the partner model.}
	\label{t8}
	\centering
	\setlength{\tabcolsep}{5pt} 
	\small
	\begin{tabular}{lr*{4}{c}}
		\toprule
		\textbf{Method} & \textbf{W} & \textbf{MR} & \textbf{FPS} & \textbf{SUR} & \textbf{DMR}\\
		\toprule
		RPN & - & 15.2 & 10 & - & -\\
		CompACT-Deep & - &  12.0 & 3 & - & -\\
		\toprule
		RPN-CompACT-A & 4.25 & 12.5 & 5 & 67\% & 0.5\\
		RPN-CompACT-F & 3 & 13.1 & 7 & 133\% & 1.1\\ 
		\bottomrule
	\end{tabular}
\end{table}
Experimental results are reported in Table~\ref{t8}. With RPN as the basic model, AF achieves satisfying speed-up ratio while maintaining acceptable miss rate. With a sampling weight of 4.25, the accurate mode is 67\% faster than the original CompACT-Deep model, at a cost of 0.5 higher MR. When the weight drops to 3, RPN-CompACT-F has 2.0 points lower miss rate than RPN at a comparable running speed. These experiments also show that the basic model can be used as an instance generator.

\section{Conclusions}

We presented adaptive feeding (AF), a simple but effective method to combine existing object detectors for speed or accuracy gains. Given one input image, AF makes a choice on either the fast (but less accurate) or the accurate (but slow) detector should be applied. Hence, AF can achieve fast and accurate detection simultaneously. The other advantage of AF is that it needs no additional training data and the training time is negligible. By combining different pairs of models, we reported state-of-the-art results on the PASCAL VOC, MS COCO and Caltech Pedestrian datasets when detection speed and accuracy are both taken into account.

Though we used pairs of models (one basic and one partner model) throughout this paper, we believe AF can be used with combinations of more than two region-based ConvNet detectors. For example, a triplet combination can adds an extra model, which is more accurate but slower than those models in our experiments~\cite{iccv2015mrcnn,yang2016craft}, to further improve the detection accuracy without losing AF's speed benefits.

\section*{Acknowledgements}
This work was supported in part by the National Natural Science Foundation of China under Grant No. 61422203.

{\small
	\bibliographystyle{ieee}
	\bibliography{egbib}
}

\end{document}